\documentclass[journal]{IEEEtran}
\usepackage{amsmath,amsfonts}

\usepackage{array}
\usepackage[caption=false,font=normalsize,labelfont=sf,textfont=sf]{subfig}
\usepackage{textcomp}
\usepackage{stfloats}
\usepackage{url}
\usepackage{verbatim}
\usepackage{graphicx}
\usepackage{cite}
\usepackage{graphicx} 
\usepackage{xcolor} 
\usepackage{amssymb}

\usepackage{colortbl}
\usepackage{booktabs}
\usepackage{tabularx}
\usepackage{multirow}
\usepackage{adjustbox}

\usepackage{amsfonts}
\usepackage{algorithm}
\usepackage{algorithmicx}
\usepackage{array}

\makeatletter
\def\blfootnote{\gdef\@thefnmark{}\@footnotetext}
\let\NAT@parse\undefined
\makeatother
\usepackage[hidelinks]{hyperref} 
\usepackage[nameinlink,capitalise]{cleveref}
\usepackage{url}

\begin{document}

\title{MambaFlow: A Novel and Flow-guided State Space Model for Scene Flow Estimation}

\author{Jiehao Luo, 
        Jintao Cheng, 
        Xiaoyu Tang,~\IEEEmembership{Member,~IEEE, 
        Qingwen Zhang, 
        Bohuan Xue, 
        Rui Fan,~\IEEEmembership{Senior Member,~IEEE}
        }
        
\thanks{Manuscript created November 2024; This work was partially supported by the National Natural Science Foundation of China under Grants 62473288 and 62233013, Guangdong Basic and Applied Basic Research Foundation (2024A1515012126), the Science and Technology Commission of Shanghai Municipal under Grant 22511104500, the Fundamental Research Funds for the Central Universities, and Xiaomi Young Talents Program. (\textit{Corresponding author: Xiaoyu Tang.})}
\thanks{Jiehao Luo and Bohuan Xue are with the School of Data Science and Engineering, and Xingzhi College, South China Normal University, Shanwei 516600, China. (e-mail: {\tt\small 20228132034@m.scnu.edu.cn}; {\tt\small bxueaa@connect.ust.hk})}

\thanks{Jintao Cheng and Xiaoyu Tang are with the School of Electronics and Information Engineering, and Xingzhi College, South China Normal University, Foshan 528225, China. (e-mail: {\tt\small 20172332035@m.scnu.edu.cn}; {\tt\small tangxy@scnu.edu.cn})}

\thanks{Qingwen Zhang is with the Division of Robotics, Perception, and Learning (RPL), KTH Royal Institute of Technology, Stockholm SE-114 28, Sweden. (e-mail: {\tt\small qingwen@kth.se})}

\thanks{Rui Fan is with the College of Electronics \& Information Engineering, Shanghai Research Institute for Intelligent Autonomous Systems, the State Key Laboratory of Intelligent Autonomous Systems, and Frontiers Science Center for Intelligent Autonomous Systems, Tongji University, Shanghai 201804, China. (e-mail: {\tt\small rfan@tongji.edu.cn})}
\thanks{Jiehao Luo and Jintao Cheng contribute equally to this work.}
}

\markboth{Journal of \LaTeX\ Class Files,~Vol.~14, No.~8, August~2021}%
{Shell \MakeLowercase{\textit{et al.}}: A Sample Article Using IEEEtran.cls for IEEE Journals}


\maketitle

\begin{abstract}
Scene flow estimation aims to predict 3D motion from consecutive point cloud frames, which is of great interest in autonomous driving field. 
Existing methods face challenges such as insufficient spatio-temporal modeling and inherent loss of fine-grained feature during voxelization. 
However, the success of Mamba, a representative state space model (SSM) that enables global modeling with linear complexity, provides a promising solution. 
In this paper, we propose MambaFlow, a novel scene flow estimation network with a mamba-based decoder. It enables deep interaction and coupling of spatio-temporal features using a well-designed  backbone. Innovatively, we steer the global attention modeling of voxel-based features with point offset information using an efficient Mamba-based decoder, learning voxel-to-point patterns that  are used to devoxelize shared voxel representations into point-wise features. 
To further enhance the model's generalization capabilities across diverse scenarios, we propose a novel scene-adaptive loss function that automatically adapts to different motion patterns.
Extensive experiments on the Argoverse 2 benchmark demonstrate that MambaFlow achieves state-of-the-art performance with real-time inference speed among existing works, enabling accurate flow estimation in real-world urban scenarios. 
The code is available at \color{blue}\url{https://github.com/SCNU-RISLAB/MambaFlow}.
\end{abstract}


\section{Introduction}

\begin{figure}
\includegraphics[width=0.49\textwidth]{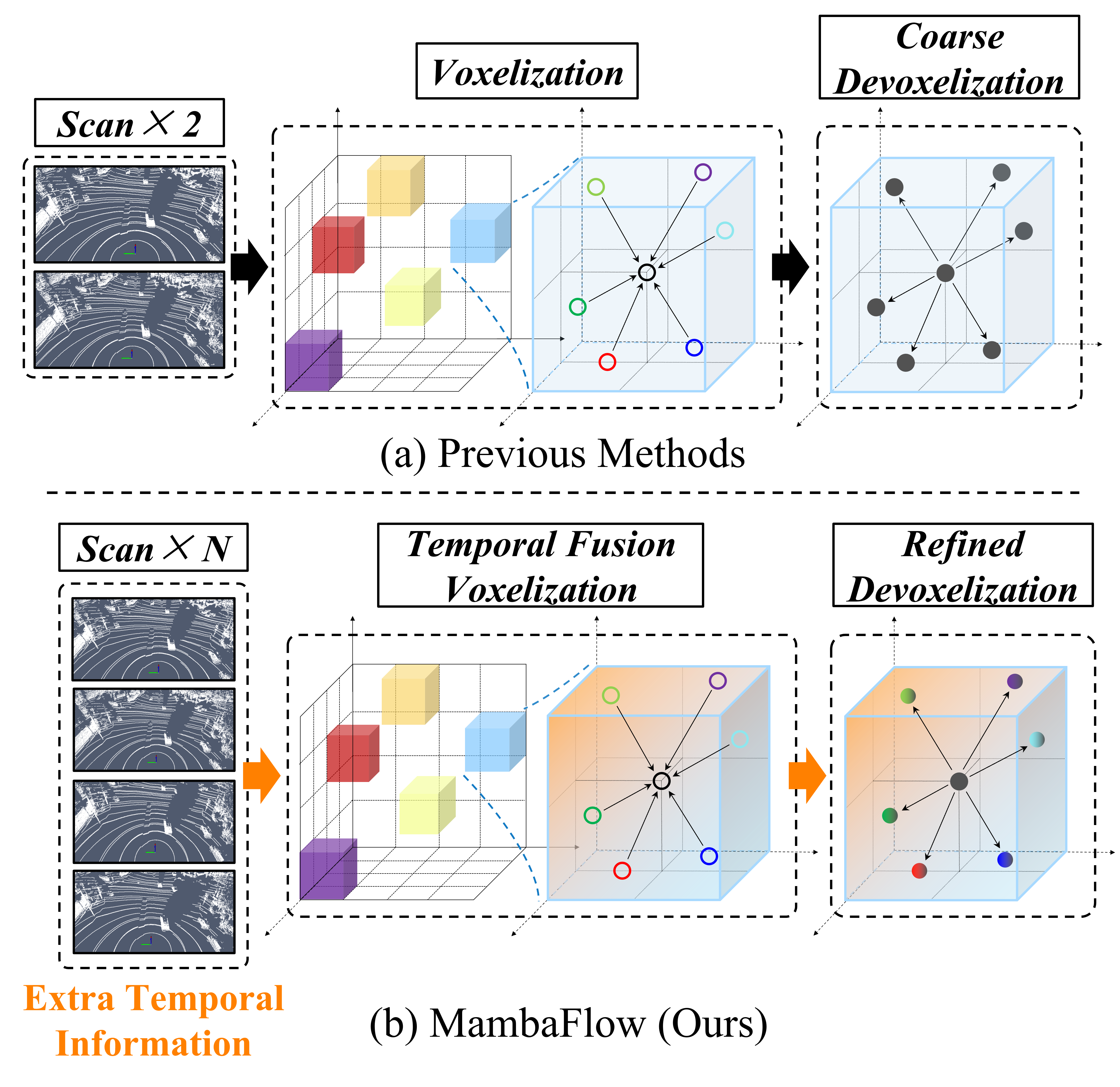}
\caption{\textbf{Comparison of voxelization and devoxelization for scene flow estimation.} (a) Previous methods typically use two consecutive frames with coarse devoxelization that assigns identical features to points in the same voxel, causing inherent feature loss. (b) Our MambaFlow leverages N consecutive scans for richer temporal information, with refined devoxelization that learns distinct voxel-to-point patterns for fine-grained feature representation.}
\label{fig:abstract}
\end{figure}

\IEEEPARstart{S}{cene} flow is a vector field that describes the motion of 3D points between two frames, representing the 3D counterpart of optical flow \cite{10258386} in 2D images. 
The scene flow estimation task takes consecutive point cloud frames from sensors as input and output motion vectors for each point, providing autonomous perception systems with essential motion information for accurate environmental perception and decision-making \cite{10458341,10570336}. Moreover, real-time scene flow estimation is also well regarded as a promising solution for enhancing the performance of downstream tasks such as 3D object detection \cite{10265176}, depth estimation \cite{10680561} and moving object segmentation \cite{10640268}, thus largely enhancing the overall intelligence of autonomous systems.

Existing scene flow estimation methods have made progress but still face several challenges. Current approaches \cite{li2021neural,li2023fast,lin2024icp,zhang2024seflow,Jund_Sweeney_Abdo_Chen_Shlens_2022,zhang2024deflow,khatri2025can} either directly concatenate consecutive point cloud frames for implicit temporal features extraction, leading to insufficient utilization of temporal information, or pursue efficiency by projecting point clouds into 2D pseudo-images, resulting in spatial feature loss. Jund et al. \cite{Jund_Sweeney_Abdo_Chen_Shlens_2022} proposed an end-to-end learning framework that concatenates consecutive point cloud frames and employs a flow embedding layer for feature extraction. Zhang et al. \cite{zhang2024deflow} drew inspiration from optical flow approaches and introduced a Gated Recurrent Unit (GRU) decoder for point cloud feature refinement. While these methods partially alleviate the feature loss, their reliance on only two consecutive frames for temporal information consistently limits model performance.

Temporal information plays a dominant role in scene flow estimation. This is effectively demonstrated by recent work Flow4D \cite{kim2024flow4d}, which leverages more than two consecutive frames to achieve significantly improved performance on the Argoverse 2 benchmark. To reduce computational complexity, Flow4D decomposes 4D convolution into parallel temporal and spatial branches with 1D and 3D convolutions respectively. While this decomposition achieves computational efficiency, it sacrifices the holistic modeling of spatio-temporal correlations. The assumption that temporal and spatial features can be processed independently ignores their inherent coupling in motion dynamics. Moreover, since temporal and spatial information characterizes fundamentally different aspects of motion dynamics and geometric structures, such simplified processing prevents effective feature interaction and temporal guidance.

To address the above challenges, we propose a novel multi-branch architecture for feature extraction that achieves deep coupling of spatio-temporal features. Specifically, our method incorporates a learnable fusion mechanism that dynamically weights temporal and spatial features while prioritizing temporal cues. This enables effective modeling of complex motion patterns across diverse scenarios, significantly enhancing the learning of motion dynamics from point clouds compared to existing methods.

Although our spatio-temporal coupling approach enables comprehensive feature extraction, the adopted voxel-based encoding introduces inherent feature loss, as points within the same voxel grid become indistinguishable. Recent works \cite{wu2024point, 10504607} demonstrate that transformer architectures can effectively learn point-wise correlations through global attention modeling. However, their quadratic inference complexity challenges real-time requirements. Studies on SSM \cite{ssm2,ssm3,ssm5} suggest a promising alternative with linear-complexity global attention modeling. Drawing inspiration from recent advances in SSM for point cloud processing \cite{mamba3d,liang2024pointmamba,mambamos}, we propose a Mamba-based decoder that steers global attention modeling of voxel-based features through point offset information. It achieves refined devoxelization by adaptively learning voxel-to-point patterns, which enables effective transformation from shared voxel representations to point-wise features, representing our core contribution to scene flow estimation.


As observed in \cite{zhang2024deflow,Jund_Sweeney_Abdo_Chen_Shlens_2022}, scene flow estimation methods often exhibit limited generalization in autonomous driving scenarios  due to the severe imbalance between dynamic and static point distributions, where over 90\% of the point cloud exhibits zero displacement. Drawing insights from self-supervised approaches \cite{Jund_Sweeney_Abdo_Chen_Shlens_2022, zhang2024deflow}, we propose a scene-adaptive loss function that leverages motion statistics derived from point displacement distributions.

Through extensive experiments, we demonstrate that our proposed feature extraction architecture integrated with the Mamba-based decoder, guided by scene-adaptive loss supervision, significantly enhances the model's dynamics-awareness and scene adaptability, achieving state-of-the-art performance on the Argoverse 2 benchmark among published works while maintaining real-time inference speed of 17.30 FPS.
The main contributions of this paper are:
\begin{enumerate}
   \item We propose MambaFlow, a novel SSM-based architecture for scene flow estimation that addresses voxelization feature loss through voxel-to-point pattern learning. To our knowledge, this represents the first application of state space modeling to scene flow estimation.
   
   \item We propose a multi-branch backbone for deep coupling of spatio-temporal features, with an adaptive fusion strategy that prioritizes temporal information for complex motion patterns.

   \item We propose a scene-adaptive loss function that leverages point displacement distributions to automatically distinguish between static and dynamic points without empirical thresholds.
\end{enumerate}

We provide our code at \color{blue}\url{https://github.com/SCNU-RISLAB/MambaFlow}\color{black}. 

\section{Related Work}

\subsection{Scene Flow Estimation Methods}
Scene flow estimation in autonomous driving, while sharing similarities with object registration methods like DifFlow3D \cite{liu2024difflow3d} and \cite{wang20233d} that achieve millimeter-level precision on small-scale datasets like ShapeNet \cite{Chang_Funkhouser_Guibas_Hanrahan_Huang_Li_Savarese_Savva_Song_Su_etal._2015} and FlyingThings3D \cite{mayer2016large}, faces unique challenges when processing scenes from datasets like Argoverse 2 \cite{argoverse2} and Waymo \cite{waymo} containing 80k-177k points per frame, where substantial downsampling compromises practicality \cite{Jund_Sweeney_Abdo_Chen_Shlens_2022}. For such large-scale data, voxel-based encoding with efficient feature refinement emerges as the preferred approach to balance efficiency and feature preservation.

Many existing methods adopt flow embedding-based approaches \cite{liu2019flownet3d,zhang2024deflow,Jund_Sweeney_Abdo_Chen_Shlens_2022,lin2024icp}, establishing soft correspondences by concatenating spatial features from consecutive frames. However, this indirect feature correlation with two-frame input has been increasingly recognized as suboptimal for capturing temporal dynamics. Following successful practices in 3D detection \cite{vedder2022sparse}, scene flow estimation has evolved towards multi-frame approaches. For example, Flow4D \cite{kim2024flow4d} processes five frames for more robust estimation. Most recently, EulerFlow \cite{vedder2024scene} reformulates scene flow as a continuous space-time PDE. However, to effectively exploit temporal dynamics while avoiding the computational complexity of continuous streams, we follow Flow4D's framework by adopting five consecutive frames as input with voxel-based encoding.

Recently, self-supervised methods have gained popularity due to the difficulty of obtaining scene flow ground truth. SeFlow \cite{zhang2024seflow} addresses key challenges through efficient dynamic classification and internal cluster consistency, while ICP-Flow \cite{lin2024icp} incorporates rigid-motion assumptions via histogram-based initialization and ICP alignment. However, precise motion estimation remains critical for autonomous driving safety, and state-of-the-art self-supervised methods like EulerFlow still show limitations in static background estimation, which could be catastrophic for autonomous driving systems. Although reducing annotation costs is appealing, the investment in labeled datasets is justified and necessary given the safety-critical nature of autonomous driving. We maintain a supervised architecture while incorporating insights from self-supervised approaches and proposed a scene-adaptive loss function, which automatically extracts motion statistics from velocity histograms for better generalization without empirical thresholds.

\subsection{State Space Model}
SSMs \cite{ssm2,ssm3,ssm5} have gained attention as an efficient alternative to attention mechanisms and Transformer architectures, particularly for capturing long-term dependencies through hidden states. The Structured State Space Sequence (S4) \cite{ssm2} efficiently models long-range dependencies through diagonal parameterization, addressing computational bottlenecks in attention-based approaches. Building upon S4, researchers developed enhanced architectures such as S5 \cite{ssm3} and H3 \cite{ssm5}. Notably, Mamba \cite{gu2023mamba} represents a significant breakthrough by introducing a data-dependent selective scan mechanism and integrating hardware-aware parallel scanning algorithms, establishing a novel paradigm distinct from CNNs and Transformers while maintaining linear computational complexity.

The linear complexity capability of Mamba has inspired extensive exploration in both vision \cite{visionmamba,efficientvmamba} and point cloud processing \cite{mamba3d,liang2024pointmamba,mambamos} domains. In 3D point cloud domain, recent works have demonstrated significant advances in adapting Mamba for various tasks. Mamba3D \cite{mamba3d} introduced local norm pooling for geometric features and a bidirectional SSM design for enhanced both local and global feature representation. PointMamba \cite{liang2024pointmamba} is one of the pioneers in applying state space models to point cloud analysis by utilizing space-filling curves for point tokenization with a non-hierarchical Mamba encoder, establishing a simple yet effective baseline for 3D vision applications. 
MambaMos \cite{mambamos} further explored SSM's potential in point cloud sequence modeling by adapting Mamba's selective scan mechanism for motion understanding, demonstrating that SSM's strong contextual modeling capabilities are particularly effective for capturing temporal correlations in moving object segmentation. These successes in achieving linear complexity while maintaining robust feature learning suggest promising potential for achieving fine-grained devoxelization in scene flow estimation.

Building upon these insights, we try to integrate the Mamba architecture into the the scene for estimation network for to maintains real-time performance while avoiding the quadratic complexity of transformer-based approaches.

\section{Problem Statement}
\label{sec:scene_flow_modeling}
Consider two consecutive point cloud frames $\mathcal{P}_t$ and $\mathcal{P}_{t+1}$ acquired at time instants $t$ and $t+1$, with vehicle ego-motion transformation matrix $\mathbf{T}_{t,t+1}$. 
The scene flow estimation task aims to predict the motion vector $\hat{\mathbf{M}}_{t,t+1}(p) = (\Delta x, \Delta y, \Delta z)^\mathrm{T}$ for each point $p$ in $\mathcal{P}_t$.
We employ the End Point Error (EPE) as the evaluation metric, as defined in~\cref{eq:epe}:
\begin{equation}
\mathrm{EPE}(p) = \|\hat{\mathbf{M}}(p) - \mathbf{M}_{gt}(p)\|_2
\label{eq:epe}
\end{equation}
where $\hat{\mathbf{M}}(p)$ and $\mathbf{M}_{gt}(p)$ denote the predicted and ground truth scene flow, respectively. 

The core objective of scene flow estimation is to minimize the average EPE, which can be expressed as~\cref{eq:aepe}:
\begin{equation}
\min \frac{1}{|\mathcal{P}_t|}\sum_{p\in \mathcal{P}_t}{\|\mathbf{\hat{M}}\left( p \right) -\mathbf{M}_{gt}\left( p \right) \|_2}
\label{eq:aepe}
\end{equation}
where $|\mathcal{P}_t|$ denotes the number of points in $\mathcal{P}_t$. A method must address both dynamic objects in static environments and global scene changes induced by ego-motion. Such challenges necessitate effective integration of local features and global context for accurate 3D motion pattern capture.

\section{Methodology}


\subsection{Preliminaries}
\label{sec:state_space_models}



Mamba \cite{gu2023mamba} is a sequence modeling framework based on SSM that introduces a selective scan mechanism to model complex state spaces through time-varying characteristics. Through a time scale parameter $\Delta$, it makes the state transition matrix $\mathbf{A}$ and input projection matrix $\mathbf{B}$ input-dependent for selective feature filtering. The continuous system is discretized via Zero-Order Hold, which can be expressed as~\cref{eq:zoh}:

\begin{equation}
\begin{gathered}
\mathbf{\bar{A}} = e^{\Delta \mathbf{A}} \\
\mathbf{\bar{B}} = (e^{\Delta \mathbf{A}} - \mathbf{I}) \mathbf{A}^{-1} \Delta \mathbf{B}
\end{gathered}
\label{eq:zoh}
\end{equation}

After discretization, the linear ordinary differential equations representing an SSM can be rewritten as~\cref{eq:rewrite}:

\begin{equation}
\begin{gathered}
h_{t} = \mathbf{\bar{A}}h_{t-1} + \mathbf{\bar{B}}x_t \\
y_t = \mathbf{C}h_{t}
\end{gathered}
\label{eq:rewrite}
\end{equation}
where $h_{t}$ denotes the hidden state, $x_t$ and $y_t$ denote the input and output sequences respectively, and $\mathbf{C}$ is output projection matrix.



\subsection{MambaFlow Pipeline}
\label{sec:mambaflow_architecture}

\subsubsection{Overall Architecture}

\begin{figure*}[t!]
    \includegraphics[width=1\textwidth]{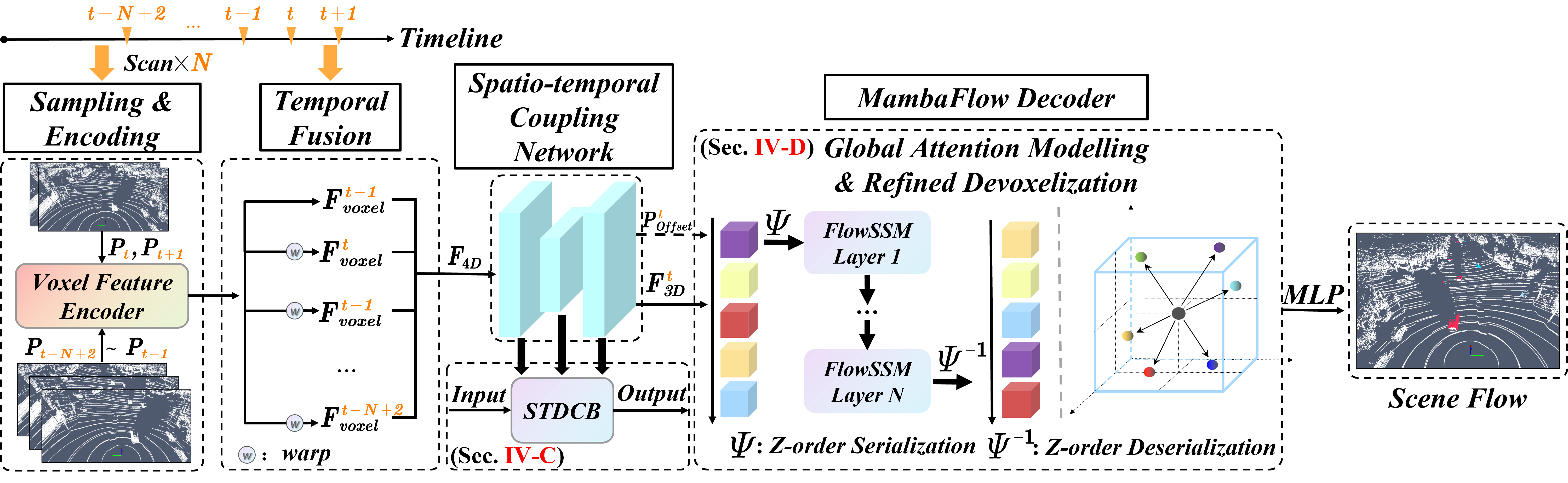}
    \caption{\textbf{Overall architecture of MambaFlow.} The network first voxelizes and encodes five consecutive scans, forming 4D features by concatenating 3D voxel representations along the temporal dimension. These features are processed by our spatio-temporal coupling network for multi-scale feature learning. The decoder then learns voxel-to-point patterns through cascaded FlowSSM layers, enabling point-wise feature differentiation within the same voxel and generating the scene flow through an MLP layer.}
    \label{fig:overview}
\end{figure*}

As shown in \Cref{fig:overview}, MambaFlow adopts an end-to-end architecture and generates scene flow estimation through three main stages: Spatio-temporal sampling and encoding, Spatio-temporal Deep Coupling Network, and MambaFlow Decoder.

In the Spatio-temporal Sampling and Encoding stage, N consecutive LiDAR scans are first transformed into 3D voxel representations through a voxel feature encoder, with all frames warped to the coordinate system of the time step $t+1$. The subsequent temporal fusion stage concatenates these 3D voxel representations along the temporal dimension, generating a 4D spatio-temporal feature tensor. 

Next, the 4D tensor is processed by the Spatio-temporal Coupling Network. This network adopts a U-Net architecture with stacked Spatio-temporal Deep Coupling Block, which progressively learns multi-scale feature representations through deep hierarchical modeling. The network consists of a five-level encoder with stacking depths [2,2,2,2,2] and a four-level decoder with stacking depths [1,1,1,1], coupling multi-scale contextual information through sparse convolution operations. The fused features are residually combined with the input features to enhance feature representation.

For the devoxelization stage, the decoder is used to processes the extracted 3D voxel features from time step $t+1$. The voxel features are first serialized into sequences following space-filling curves. Through multiple cascaded FlowSSM modules, which incorporate discretized point offset information into state space for the global modeling of voxel-wise features, the decoder progressively refines voxel-wise features to point-wise features. Finally, the refined feature sequence is deserialized and fed into an MLP head to generate point-wise scene flow estimation.

\subsubsection{Input Representation and Voxelization}
To balance prediction performance and computational efficiency, we follow Flow4D~\cite{kim2024flow4d} by sampling five consecutive point cloud frames as input.

Given consecutive point cloud frames $\mathcal{P}_\tau$ that can be represented as~\cref{eq:seq}:
\begin{equation}
\begin{gathered}
\mathcal{P}_\tau = \{p_i \in \mathbb{R}^3\}_{i=0}^{N_\tau-1}, \\
\tau \in \{t-3, t-2, t-1, t, t+1\}
\end{gathered}
\label{eq:seq}
\end{equation}
where $p_i = (x_i, y_i, z_i)^{\mathrm{T}}$ denotes the point coordinates and $N_\tau$ denotes the number of points at time step $\tau$.


Following previous studies, we first warp all point clouds to the perspective of time step $t+1$ using the known pose transformation matrices. For each transformed point cloud, we first employ an $\mathrm{MLP}$ to extract point-wise features $\boldsymbol{F}_{point}^{\tau}$, followed by a voxel feature encoder that aggregates these features into voxel-wise features $\mathbf{F}_{voxel}^{\tau}$. To form spatio-temporal representations, we extend $\boldsymbol{F}_{voxel}^{\tau}$ with a temporal dimension and concatenate them along the time axis, yielding a 4D voxel tensor $\boldsymbol{F}_{4D}$ that encodes both spatial and temporal information.




\subsubsection{Serialization}
Since our SSM-based decoder processes sequential data, we adopt Z-order space-filling curves as our serialization strategy inspired by \cite{wu2024point}. We define a mapping function $\Psi$ that projects 3D point cloud $\mathcal{P}_{o}$ to sequence $\mathcal{P}_{o}^{\prime}$ while preserving spatial locality. The serialization and deserialization process can be expressed as~\cref{eq:serialisation}:

\begin{equation}
\begin{aligned}
\mathcal{P}_{o}^{\prime} &= \Psi(\mathcal{P}_{o})\\
\mathcal{P}_{o} &= \Psi^{-1}(\mathcal{P}_{o}^{\prime})
\end{aligned}
\label{eq:serialisation}
\end{equation}

\subsection{Spatio-temporal Deep Coupling Block}
\label{sec:stdcb}

\begin{figure*}[t!]
    \includegraphics[width=1.01\textwidth]{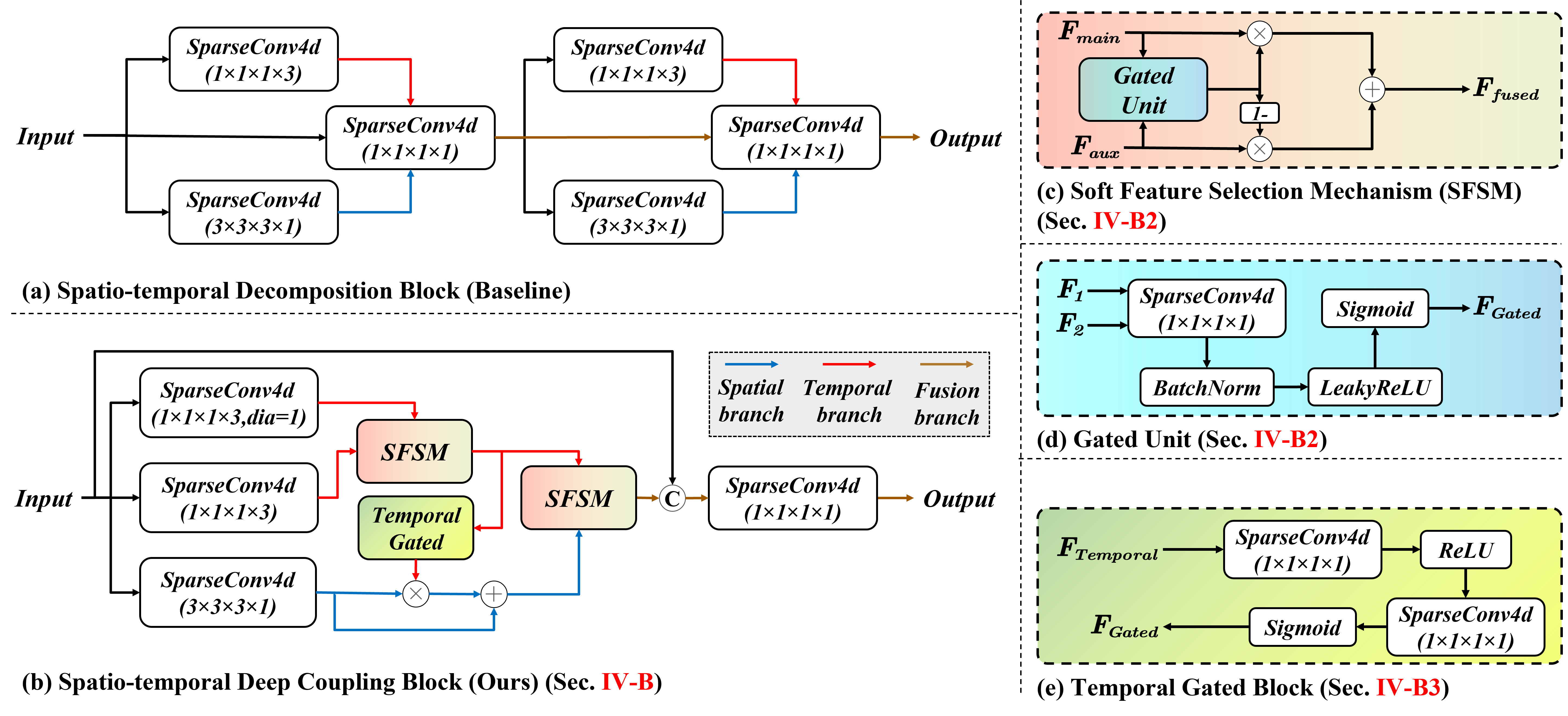}
    \caption{\textbf{Architecture of the Spatio-temporal Deep Coupling Block.} (a) The baseline Spatio-temporal Decomposition Block \cite{kim2024flow4d} processes features through repeated convolutions at each stage. (b) Our proposed Spatio-temporal Deep Coupling Block achieves more efficient feature extraction by removing redundant convolutions modules and introducing a cross-timestep branch. The right panel (c)-(e) shows the detailed structures of Soft Feature Selection Mechanism and gating mechanisms used in Spatio-temporal Deep Coupling Block.}
    \label{fig:stdcb}
\end{figure*}
\subsubsection{Feature Extraction Stage}

As shown in \Cref{fig:stdcb}, the proposed spatio-temporal deep coupling block (STDCB) consists of multi sparse convolutions branches, incorporating a cascaded Soft Feature Selection Mechanism to achieve adaptive feature interaction. Specifically, STDCB first sparsifies $\boldsymbol{F}_{4D}$ to obtain a sparse 4D tensor $\boldsymbol{F}_\text{sparse}$, which is then processed by three parallel branches. Drawing inspiration from the design of Flow4D \cite{kim2024flow4d}, we adopt a convolution with kernel size $(\textit{3}\times\textit{3}\times\textit{3}\times\textit{1})$ to extract geometric structures, and two convolution with kernel size $(\textit{1}\times\textit{1}\times\textit{1}\times\textit{3})$ to capture local motion patterns and cross-timestep dependencies through different receptive fields. The process of initially feature extraction can be formulated as~\cref{eq:feature_extraction}:

\begin{equation}
\begin{aligned}
\boldsymbol{F}_{spatial} &= \Phi_{\textit{3}\times\textit{3}\times\textit{3}\times\textit{1}}(\boldsymbol{F}_{sparse}) \\
\boldsymbol{F}_{temporal} &= \Phi_{\textit{1}\times\textit{1}\times\textit{1}\times\textit{3}}(\boldsymbol{F}_{sparse}) \\
\boldsymbol{F}_{temporal}^{ct} &= \Phi_{\textit{1}\times\textit{1}\times\textit{1}\times\textit{3},\textit{dilation=1}}(\boldsymbol{F}_{sparse})
\end{aligned}
\label{eq:feature_extraction}
\end{equation}
where $\boldsymbol{F}_{spatial}$, $\boldsymbol{F}_{temporal}$ and $\boldsymbol{F}_{temporal}^{ct}$ denote the features extracted from spatial, local temporal, and cross-timestep temporal branches respectively. $\Phi$ denotes the convolution kernel. For the cross-timestep, we employ a dilated convolution operation where a dilation factor of 1 is applied to expand the receptive field.

The decomposed design significantly reduces computational complexity compared to full 4D convolutions while preserving the ability to capture both spatial and temporal information.




\subsubsection{Soft Feature Selection Mechanism}
\label{sfsm}
Given two feature tensors as input, Soft Feature Selection Mechanism designates one as the main branch $\boldsymbol{F}_{main}$ and the other as the auxiliary branch $\boldsymbol{F}_{aux}$ according to their roles in the specific task. Their relative importance is then computed through a Gated Unit. The attention weights can be formulated as~\cref{eq:attention}:
\begin{equation}
\mathbf{\alpha} = \sigma(\mathrm{LR}(\mathrm{BN}(\Phi_{p}(\mathrm{Concat}(\boldsymbol{F}_{main}, \boldsymbol{F}_{aux})))))
\label{eq:attention}
\end{equation}
where $\sigma$ denotes the sigmoid activation, $\mathrm{LR}$ denotes the LeakyReLU, $\mathrm{BN}$ denotes the batch normalization operation, $\Phi_{p}$ denotes the point-wise convolution, and $\mathrm{Concat}$ denotes feature concatenation along the channel dimension.

The attention weights are then used to adaptively combine the two branches, where $\mathbf{\alpha}$ is multiplied with the main branch and $(1-\mathbf{\alpha})$ with the auxiliary branch. This weighting scheme allows flexible control of feature importance: when $\mathbf{\alpha}$ is large, the main branch features are emphasized, while smaller $\mathbf{\alpha}$ values give more weight to the complementary features from the auxiliary branch. 


\subsubsection{Temporal Gated Block}
\label{tgb}
After obtaining features from three parallel branches, we first fuse temporal features by employing Soft Feature Selection Mechanism (SFSM) with consecutive temporal features as the main branch and cross-timestep temporal features as the auxiliary branch to complement cross-step temporal information, which can be formulated as~\cref{eq:temp_fusion}:
\begin{equation}
\boldsymbol{F}_{temporal}^{\prime} = \mathrm{SFSM}(\boldsymbol{F}_{temporal}, \boldsymbol{F}_{temporal}^{ct})
\label{eq:temp_fusion}
\end{equation}

The fused temporal features then guide spatial feature learning through the Temporal Gated Block, where the attention weights $\mathbf{\beta}$ are computed as~\cref{eq:temp_attention}:
\begin{equation}
\mathbf{\beta} = \sigma(\Phi_p(\mathrm{ReLU}(\Phi_p(\boldsymbol{F}_{temporal}^\prime))))
\label{eq:temp_attention}
\end{equation}

The spatial features are modulated by temporal attention through gating and residual connection, which can be formulated as~\cref{eq:spatial_mod}:
\begin{equation}
\boldsymbol{F}_{spatial}^{\prime} = \boldsymbol{F}_{spatial} \odot (1 + \mathbf{\beta})
\label{eq:spatial_mod}
\end{equation}
where $\odot$ denotes element-wise multiplication.

Finally, $\boldsymbol{F}_{spatial}^{\prime}$ is adaptively combined with $\boldsymbol{F}_{temporal}^{\prime}$ using SFSM, followed by a residual connection with $\boldsymbol{F}_{sparse}$ and a point-wise convolution to fuse the features. The final output can be formulated as~\cref{eq:final_output}:
\begin{equation}
\boldsymbol{F}_{sparse}^{\prime} = \Phi_p(\mathrm{Concat}(\mathrm{SFSM}(\boldsymbol{F}_{temporal}^{\prime}, \boldsymbol{F}_{spatial}^{\prime});\boldsymbol{F}_{sparse}))
\label{eq:final_output}
\end{equation}


\subsection{MambaFlow Decoder}
\label{sec:mambaflow_decoder}



\begin{figure}[t!]
        \includegraphics[width=0.48\textwidth]{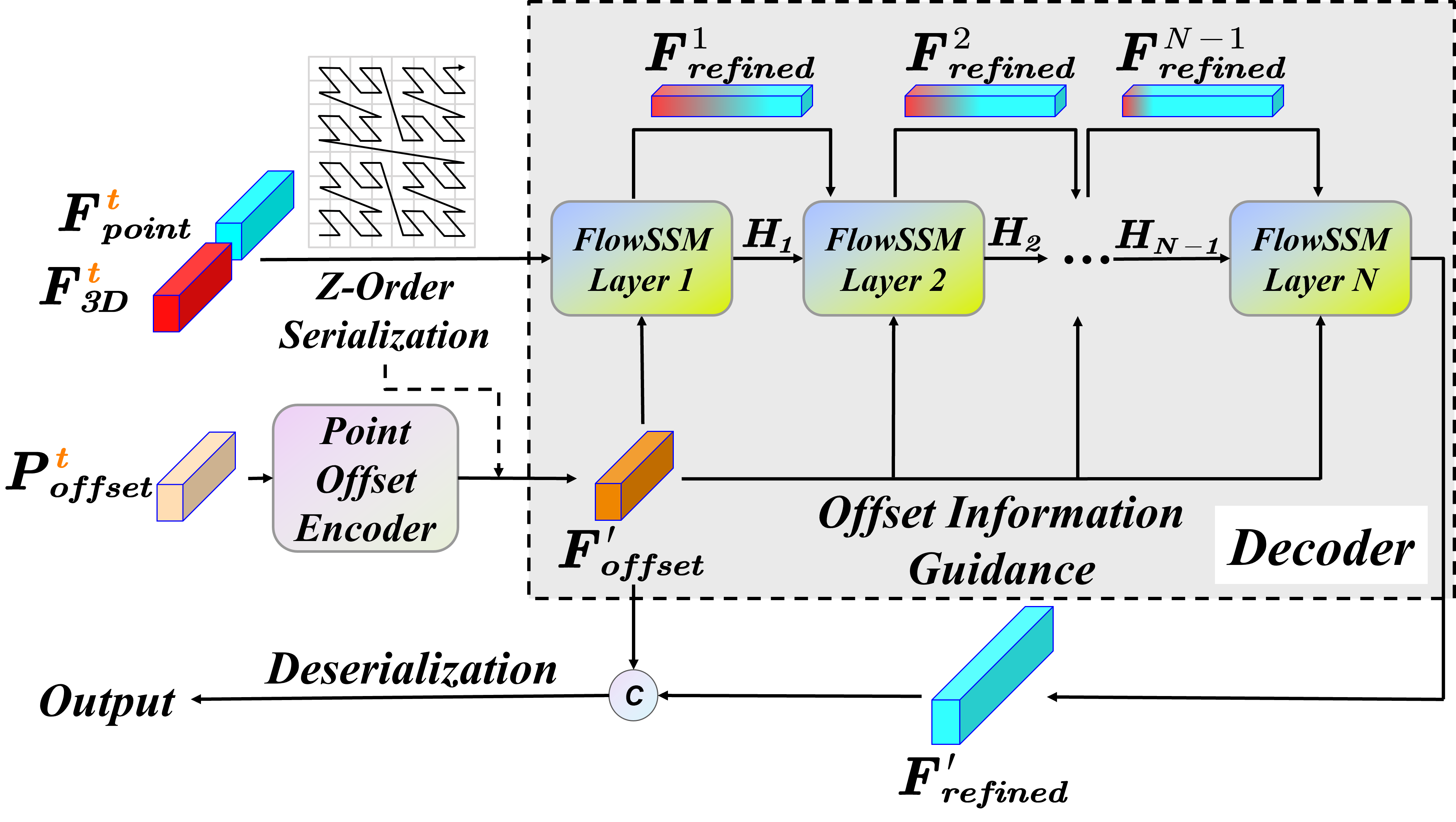}
        \caption{\textbf{MambaFlow Decoder architecture.} Point-wise features and voxel features are first serialized through Z-order space-filling curves for spatial proximity preservation. The FlowSSM module consists of N cascaded FlowSSM layers, where point offset features guide the learning of voxel-to-point patterns in each layer for refined feature reconstruction. The final output is obtained through deserialization and feature fusion with point offset information.}
        \label{fig:decoder}
\end{figure}

The overall structure of the MambaFlow Decoder is shown in \Cref{fig:decoder}. During the decoding stage, we first extract the 3D voxel features $\boldsymbol{F}_{3D}^t$ at time step $t$ from $\boldsymbol{F}_{sparse}^{\prime}$. For each point in a voxel grid, we assign the corresponding voxel features as their initial point-wise representations, denoted as $\boldsymbol{F}_{coarse}^t$, where all points in the same voxel share identical feature values.

Meanwhile, we preserve both point-wise features $\boldsymbol{F}_{point}^t$ and point offset information $\boldsymbol{P}_{offset}^t$ during encoding stage. Considering that $\boldsymbol{P}_{offset}^t$ only have 3 channels, which may cause information imbalance when directly used as decoder input, we extend their feature dimension to a matching dimension of $\boldsymbol{F}_{3D}^t$ through a point offset encoder, yielding point-wise offset features $\boldsymbol{F}_{offset}^t$. The concatenation of $\boldsymbol{F}_{3D}^t$ and $\boldsymbol{F}_{point}^t$ along the channel dimension forms the initial input features $\boldsymbol{F}_{coarse}^{0}$ to the first decoding layer, with $\boldsymbol{F}_{offset}^t$ serving as guiding information for feature refinement. 

The key component of our decoder is FlowSSM, a variant of SSM. As shown in \cref{alg:flow_ssm}, it incorporates discretized point offset information into state space for global modeling of voxel-wise features. To enable sequence modeling, we first serialize both coarse features and offset features using Z-order space-filling curves, which can be formulated as~\cref{eq:serialize}:
\begin{equation}
\begin{gathered}
\boldsymbol{F}_{coarse}^{\prime} = \Psi(\boldsymbol{F}_{coarse}^t) \\
\boldsymbol{F}_{offset}^{\prime} = \Psi(\mathbf{P}_{{offset}}^t)
\end{gathered}
\label{eq:serialize}
\end{equation}


The decoder then progressively refines these features through multiple cascaded FlowSSM layers, where each layer conditions the state space matrices on $\boldsymbol{F}_{offset}^{\prime}$ for adaptive feature refinement. After $N$ iterations, the refined feature sequence is first deserialized to obtain refined point-wise features $\boldsymbol{F}_{refined}^{\prime}$, then concatenated with $\boldsymbol{F}_{offset}^{\prime}$. Finally, the scene flow estimation $\hat{\mathbf{M}}_{t,t+1}(p)$ is generated through deserialization and an MLP head.




\subsection{Scene-adaptive Loss}
\label{sec:scene-adaptive_loss}

In scene flow estimation, supervised methods relying on manually labeled data often show limited generalization ability to real-world scenes. Inspired by the the loss function design of FastFlow3D \cite{Jund_Sweeney_Abdo_Chen_Shlens_2022}, DeFlow \cite{zhang2024deflow} divides points into three subsets \{$\mathcal{P}_{1}, \mathcal{P}_{2}, \mathcal{P}_{3}$\} using velocity thresholds of 0.4 m/s and 1.0 m/s, and computes the total loss by averaging the endpoint errors within each subset, which can be formulated as~\cref{eq:total_loss}:
\begin{equation}
\begin{aligned}
\mathcal{L} = \sum_{i=1}^3\frac{1}{|\mathcal{P}_i|} \sum_{p \in \mathcal{P}_{i}}\|\Delta \hat{\mathbf{M}}(p) - \Delta \mathbf{M}_{\text{gt}}(p)\|_2
\end{aligned}
\label{eq:loss}
\end{equation}

However, empirically determined velocity thresholds can misclassify slow-moving objects as static points. Considering the severe imbalance where over 90\% of points exhibit zero displacement in autonomous driving scenarios, we propose a scene-adaptive loss function that automatically determines the displacement threshold for effective model training.

Specifically, we first divide the displacement range $[r_{\text{min}}, r_{\text{max}}]$ of a point cloud frame into $K$ equal-width bins, where $r$ denotes the point displacement and $r_{\text{min}}$ is set to 0 considering that zero-displacement points are in the majority of all points. We define $w_j$ as the proportion of points whose displacement magnitudes fall into the $j$-th bin $b_j$, which can be formulated as~\cref{eq:bin_boundaries}:
\begin{equation}
w_j = \frac{n_j}{n} 
\label{eq:bin_boundaries}
\end{equation}
where $n_j$ denotes the number of points in $j$-th bin and $n$ is the total number of points.

We then adaptively select the first bin whose scale is below $1/K$ as the displacement threshold $\alpha$, which can be expressed as~\cref{eq:threshold_calculation}:
\begin{equation}
\alpha = \min\{j : w_j < \frac{1}{K}\}
\label{eq:threshold_calculation}
\end{equation}

\definecolor{mygreen}{RGB}{0,100,0}
\begin{algorithm}[t]
\caption{\enskip \textbf{FlowSSM Process}}\label{alg:flow_ssm}
\begin{algorithmic}
    \State \textbf{Require}: Coarse flow features $\boldsymbol{F}_{coarse}^{i-1}$: \textcolor{mygreen}{($B,N ,2C$)}
    \Statex \qquad Point offset features $\boldsymbol{F}_{offset}^{\prime}$: \textcolor{mygreen}{($B,N,C$)} 
    \Statex \qquad Hidden state $\mathbf{H}_{i-1}$: \textcolor{mygreen}{($B,D,N$)} 
    \State \textbf{Ensure}: Refined flow features $\boldsymbol{F}_{coarse}^{i}$: \textcolor{mygreen}{($B,N,C$)}
    \Statex \qquad Updated hidden state $\mathbf{H}_{i}$: \textcolor{mygreen}{($B,D,N$)}
    \Statex /* Parameterize data independent matrices */
    \State $\mathbf{A}$: \textcolor{mygreen}{($D,N$)} $\leftarrow$ Parameter
    \State $\mathbf{D}$: \textcolor{mygreen}{($D,$)} $\leftarrow$ Parameter
    \Statex /* Parameterize data dependent matrices via point offset features */
    \State $\Delta$: \textcolor{mygreen}{($B,D$)}, $\mathbf{B}$: \textcolor{mygreen}{($B,N$)}, $\mathbf{C}$: \textcolor{mygreen}{($B,N$)} $\leftarrow \text{Linear}(\boldsymbol{F}_{offset}^{\prime})$
    \Statex /* Discretize */
    \State ${\mathbf{\bar{A}}}$: \textcolor{mygreen}{($B,D,N$)} $ \leftarrow \text{Exp}(\Delta \otimes \mathbf{A})$
    \State ${\mathbf{\bar{B}}}$: \textcolor{mygreen}{($B,D,N$)} $ \leftarrow \Delta \otimes \mathbf{B}$
    \Statex /* Running SSM */
    \State $\boldsymbol{F}_{coarse}^{i}$: \textcolor{mygreen}{$(B,N,2C)$}, $\mathbf{H}^{i}$: \textcolor{mygreen}{$(B,D,N)$} \State \hfill $\leftarrow \text{SSM}({\mathbf{\bar{A}}}, {\mathbf{\bar{B}}}, \mathbf{C}, \mathbf{D})(\boldsymbol{F}_{coarse}^{i-1}, \mathbf{H}_{i-1})$
    \State \textbf{Return}: $\boldsymbol{F}_{refind}^{i}, \mathbf{H}_{i}$
\end{algorithmic}
\end{algorithm}

By setting $K=100$, we can effectively capture the inherent displacement distribution of each scene and focus the loss on the truly dynamic portion of the point cloud. With the threshold $\alpha$ indicating the bin index, we divide the point $p$ into two categories \{$\mathcal{P}_{\text{Static}}, \mathcal{P}_{\text{Dynamic}}$\} based on their displacement $r(p)$ relative to the lower bound of the $\alpha$-th bin $r_\alpha$, as defined in~\cref{eq:point_sets}:

\begin{equation}
\begin{aligned}
\mathcal{P}_{\text{Static}} &= \{p \in \mathcal{P} \mid r(p) \leq r_\alpha\} \\[1ex]
\mathcal{P}_{\text{Dynamic}} &= \{p \in \mathcal{P} \mid r(p) > r_\alpha\}
\end{aligned}
\label{eq:point_sets}
\end{equation}

The total loss function is a weighted average of these two types of losses, which can be formulated as~\cref{eq:total_loss}:
\begin{equation}
\begin{aligned}
\mathcal{L}_{\text{total}} = \frac{1}{|\mathcal{P}_{\text{Static}}|} \sum_{p \in \mathcal{P}_{\text{Static}}} \|\Delta \hat{\mathbf{M}}(p) - \Delta \mathbf{M}_{\text{gt}}(p)\|_2 \\
+ \frac{1}{|\mathcal{P}_{\text{Dynamic}}|} \sum_{p \in \mathcal{P}_{\text{Dynamic}}} \|\Delta \hat{\mathbf{M}}(p) - \Delta \mathbf{M}_{\text{gt}}(p)\|_2
\end{aligned}
\label{eq:total_loss}
\end{equation}
where $|\mathcal{P}_{\text{Static}}|$ and $|\mathcal{P}_{\text{Dynamic}}|$ denote the number of the static and dynamic point sets respectively, $\|\cdot\|_2$ denotes the L2 norm.


\section{Experiment}

\begin{table*}
\caption{\textbf{Quantitative comparison on the Argoverse 2 test set.} 'Sup.' represents supervised methods. 'FD' represents Foreground Dynamic, 'BS' represents Background Static, and 'FS' represents Foreground Static.}
\centering
\small
\begin{tabular}{@{}lccccccccccc@{}}
\toprule
\multirow{5}{*}{Method} & \multirow{5}{*}{Sup.}  & \multicolumn{4}{c}{3-way Endpoint Error ($\downarrow$)} & \multicolumn{6}{c}{Bucketed Normalized Endpoint Error ($\downarrow$)} \\
\cmidrule(lr){3-6} \cmidrule(l){7-12}
& & \multirow{3}{*}{Avg.} & \multirow{3}{*}{FD} & \multirow{3}{*}{BS} & \multirow{3}{*}{FS} & \multicolumn{5}{c}{Dynamic} & \multirow{3}{*}{\begin{tabular}[c]{@{}c@{}}Static\\mean\end{tabular}} \\
\cmidrule(lr){7-11}
& & & & & & mean & Car & \begin{tabular}[c]{@{}c@{}}Other\\vehicles\end{tabular} & \begin{tabular}[c]{@{}c@{}}Pedes-\\trian\end{tabular} & \begin{tabular}[c]{@{}c@{}}Wheeled-\\vru\end{tabular} &  \\

\midrule
NSFP  & & 0.0606 & 0.1158 & 0.0344 & 0.0316 & 0.4219 & 0.2509 & 0.3313 & 0.7225 & 0.3831 & 0.0279  \\
FastNSF  & & 0.1118 & 0.1844 & 0.0907 & 0.0814 & 0.3826 & 0.2901 & 0.4126 & 0.5002 & 0.3215 & 0.0736  \\
ICP-Flow  & & 0.0650 & 0.1369 & 0.0250 & 0.0332 & 0.3309 & 0.1945 & 0.3314 & 0.4353 & 0.3626 & 0.0271  \\
SeFlow  & & 0.0536 & 0.1323 & 0.0043 & 0.0242 & 0.3194 & 0.2178 & 0.3464 & 0.4452 & 0.2683 & 0.0148  \\
EulerFlow  & & 0.0423 & 0.0498 & 0.0526 & 0.0245 & \textbf{0.1303} & 0.0929 & \underline{0.1408} & \textbf{0.1947} & \textbf{0.0931} & 0.0253  \\
FastFlow3D  & $\checkmark$ & 0.0735 & 0.1917 & 0.0027 & 0.0262 & 0.5323 & 0.2429 & 0.3908 & 0.9818 & 0.5139 & 0.0182  \\
DeFlow  & $\checkmark$ & 0.0501 & 0.1091 & 0.0062 & 0.0352 & 0.3704 & 0.1530 & 0.3150 & 0.6615 & 0.3520 & 0.0262  \\
TrackFlow  & $\checkmark$ & 0.0473 & 0.1030 & \underline{0.0024} & 0.0365 & 0.2689 & 0.1817 & 0.3054 & 0.3581 & 0.2302 & 0.0447  \\
Flow4D  & $\checkmark$ & \underline{0.0224} & \underline{0.0494} & 0.0047 & \underline{0.0130} & 0.1454 & \underline{0.0871} & 0.1505 & \underline{0.2165} & 0.1272 & \underline{0.0106} \\
\midrule
MambaFlow (Ours) & $\checkmark$ & \textbf{0.0191}& \textbf{0.0450}& \textbf{0.0015}& \textbf{0.0108} & \underline{0.1422}& \textbf{0.0786}& \textbf{0.1399}& 0.2369& \underline{0.1136}& \textbf{0.0077}\\
\bottomrule
\end{tabular}
\label{table:comparison_test}
\end{table*}

\begin{table*}[!t]
\caption{\textbf{Quantitative comparison on the Argoverse 2 validation set.}}
\centering
\renewcommand{\arraystretch}{1.1}  
\begin{tabular}{l ccccccc}
\toprule
\multirow{3}{*}{Method} & \multicolumn{2}{c}{Bucketed Normalized EPE ($\downarrow$)} & \multicolumn{4}{c}{3-way Endpoint Error ($\downarrow$)} & \multirow{3}{*}{\begin{tabular}[c]{@{}c@{}}Dynamic IoU\\ ($\uparrow$)\end{tabular}} \\
\cmidrule(lr){2-3} \cmidrule(lr){4-7}
 & mean Dynamic & mean Static & Avg. & FD & BS & FS & \\[0.5ex]
\midrule
FastFlow3D  & 0.3975 & 0.0113 & 0.0461 & 0.1218 & 0.0028 & 0.0136 & 0.7169 \\
DeFlow  & 0.3299 & 0.0192 & 0.0446 & 0.1013 & 0.0043 & 0.0281 & 0.7505 \\
Flow4D  & 0.1775 & 0.0092 & 0.0206 & 0.0468 & 0.0033 & 0.0117 & 0.8137 \\
\midrule
MambaFlow (Ours) & \textbf{0.1620} & \textbf{0.0079} & \textbf{0.0176} & \textbf{0.0414} & \textbf{0.0018} & \textbf{0.0096} & \textbf{0.8661} \\
\bottomrule
\end{tabular}
\label{table:comparison_val}
\end{table*}

\subsection{Experiment Setups}

\subsubsection{Datasets}

The proposed method is evaluated on the Argoverse 2 dataset \cite{argoverse2}, a large-scale autonomous driving benchmark containing 1,000 diverse scenarios. Each sequence in this dataset spans 15 seconds and encompasses 30 distinct object classes in complex urban environments. The dataset provides comprehensive sensor data, including LiDAR scans, detailed 3D annotations, and HD maps, making it particularly suitable for evaluating scene flow estimation in real-world urban scenarios. 


\subsubsection{Metrics}
We employ two metrics to evaluate our method. First, we use the standard End Point Error (EPE), which measures the L2 distance between predicted and ground truth scene flow vectors. However, as EPE tends to be dominated by large objects and fails to reflect performance on smaller, safety-critical objects like pedestrians, we also adopt Bucket Normalized EPE \cite{khatri2025can}. This metric evaluates each object category separately and normalizes the error by object speed, enabling more balanced assessment across different object types and motion patterns.

\subsubsection{Implementation details}
Our model is implemented in PyTorch with a two-phase training strategy. The first phase focuses on training the backbone network with scene-adaptive loss without the decoder to learn robust feature representations. In the second phase, we integrate the decoder and fine-tune the entire architecture with a lower learning rate to preserve the pre-trained backbone knowledge while optimizing decoder performance.

We conduct distributed training on 8 NVIDIA RTX 4090 GPUs. The first phase runs for 30 epochs with an initial learning rate of 2e-4 and batch size of 5 per GPU. The second phase continues for 50 epochs with an initial learning rate of 2e-6 and batch size of 3 per GPU, while maintaining other hyperparameters.

\subsection{Scene Flow Estimation Performance}
\subsubsection{Quantitation Analysis}

We compare our proposed MambaFlow with various supervised and self-supervised scene flow methods.~\cref{table:comparison_test} presents the comparative results evaluated through the official Argoverse 2 test server. The proposed method achieves state-of-the-art performance in all EPE metrics
, demonstrating the most accurate point-level motion estimation across all published works. Specifically, MambaFlow surpasses Flow4D by significant margins in all three metrics, showing a 14.7\% improvement in average EPE, 8.9\% in foreground dynamic estimation, and 68.1\% in background static prediction.

For object-level evaluation using Bucketed Normalized EPE, our method achieves competitive performance in dynamic object estimation, particularly excelling in rigid objects such as cars and other vehicles.
MambaFlow's balanced performance across both static and dynamic scenarios, especially its superior point-level accuracy as demonstrated by 3-way EPE metrics, validates its effectiveness as a comprehensive scene flow estimation solution.

As shown in~\cref{table:comparison_val}, we further compare our method with supervised methods FastFlow3D, DeFlow, and Flow4D on the Argoverse 2 validation set. MambaFlow achieves best performance across all metrics, demonstrating an 8.7\% lower mean Dynamic Normalized EPE, 14.6\% lower 3-way mean EPE, and 6.4\% higher Dynamic IoU compared to the previous state-of-the-art Flow4D.

\subsubsection{Qualitative Analysis}

\begin{figure*}[t!]
        \includegraphics[width=1\textwidth]{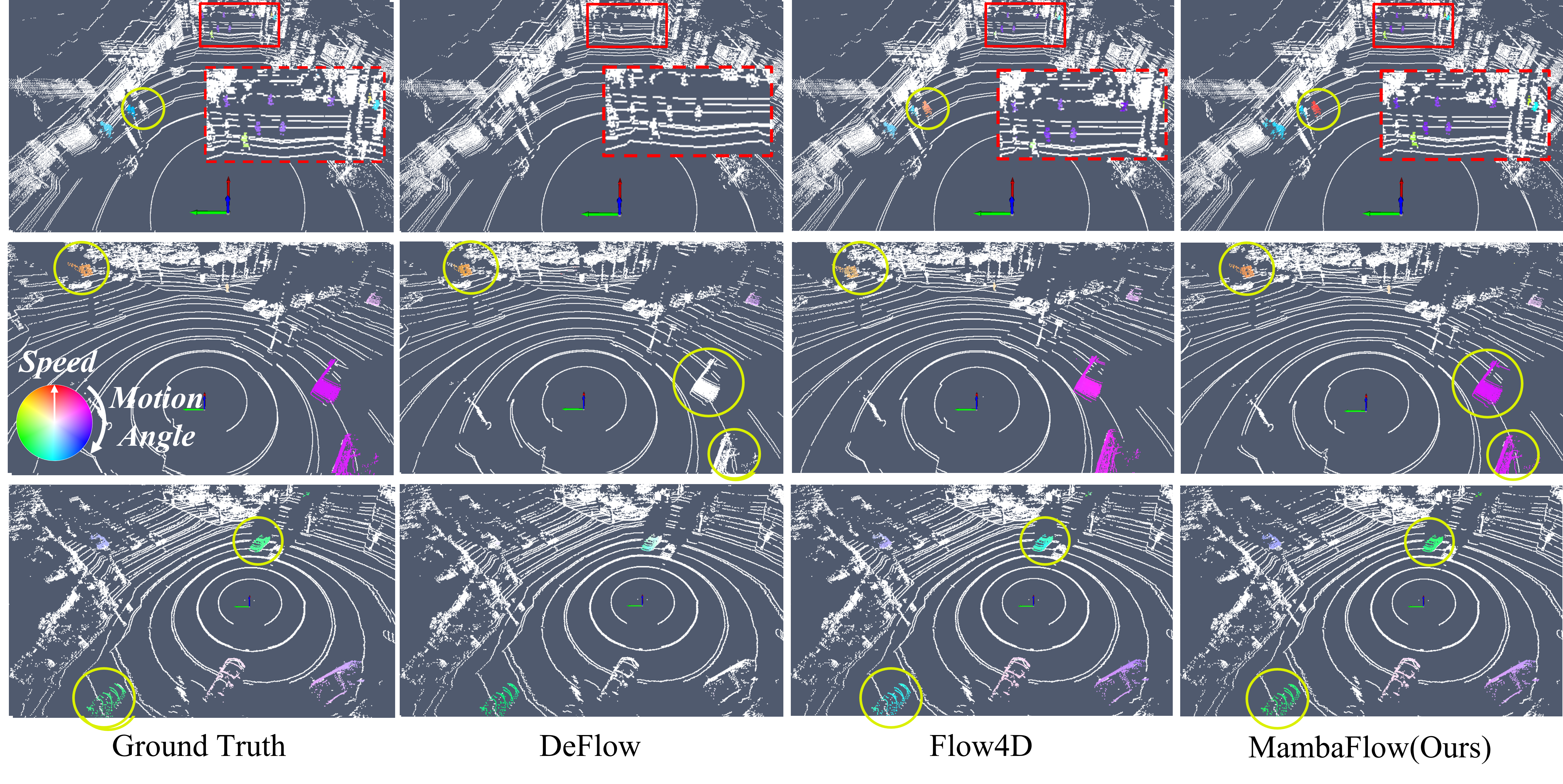}
        \caption{\textbf{Qualitative results on the Argoverse 2 validation set.} From left to right: Ground Truth, DeFlow, Flow4D, and our proposed MambaFlow. The color legend indicates both speed (shown by color intensity) and motion angle (2D), aligned with the vehicle's forward direction. The highlighted regions (yellow circles) demonstrate our method's superior performance in capturing both static and dynamic object motions, especially for challenging cases with complex motion patterns.}
        \label{fig:visualizion}
\end{figure*}

As shown in \Cref{fig:visualizion}, we visualize scene flow predictions from different methods alongside ground truth across multiple scenarios, where our method demonstrates superior discrimination between moving and static objects, producing minimal flow errors for various structures including buildings, pedestrians, and parked vehicles across multiple scenarios. This capability is crucial for reliable scene understanding in autonomous driving scenarios.

Furthermore, our method exhibits remarkable capability in object-level scene flow estimation. For instance, in the first row (yellow circles), where two pedestrians are walking in opposite directions, MambaFlow accurately captures their distinct motion patterns despite the lack of explicit ground truth annotations for this scenario. In contrast, Flow4D shows only minimal response to these opposing movements, while DeFlow fails to detect this complex interaction entirely. This demonstrates our method's superior sensitivity to fine-grained motion patterns at the object level.
\begin{table}[!t]
\caption{\textbf{Performance comparison of different component configurations.} `STDCB' represents the spatio-temporal deep coupling block, `M.F. Decoder' represents the MambaFlow Decoder and `S.A. Loss' represents the Scene-Adaptive Loss.}
\centering
\setlength{\tabcolsep}{0.6em}
\small
\renewcommand{\arraystretch}{1.1}
\begin{tabular}{l ccc cc}
\toprule
\multirow{2}{*}{Method} & \multicolumn{3}{c}{Components} & \multicolumn{2}{c}{B.N. EPE ($\downarrow$)} \\
\cmidrule(lr){2-4} \cmidrule(lr){5-6}
 & STDCB & \begin{tabular}[t]{@{}c@{}}M.F.\\Decoder\end{tabular} & \begin{tabular}[t]{@{}c@{}}S.A.\\Loss\end{tabular} & mean D. & mean S. \\
\midrule
Baseline & -  & - & - & 0.1775 & 0.0092 \\
\midrule
\multirow{3}{*}{(i)} & $\checkmark$ & & & 0.1685 & 0.0087 \\
 & & $\checkmark$ & & 0.1720 & 0.0091 \\
 & & & $\checkmark$ & 0.1744 & 0.0078 \\
\midrule
\multirow{3}{*}{(ii)} & $\checkmark$ & $\checkmark$ & & 0.1646 & 0.0090 \\
 & $\checkmark$ & & $\checkmark$ & 0.1645 & 0.0080 \\
 & & $\checkmark$ & $\checkmark$ & 0.1695 & 0.0085 \\
\midrule
MambaFlow & $\checkmark$ & $\checkmark$ & $\checkmark$ & \textbf{0.1620} & \textbf{0.0079} \\
\bottomrule
\end{tabular}
\label{table:ablation}
\end{table}

\subsection{Ablation Study}
\begin{table}[!t]
\caption{\textbf{Ablation study on FlowSSM iteration count.} '\#Itr' represents the different numbers of iterations.}
\centering
\small
\renewcommand{\arraystretch}{1.1}
\begin{tabular}{c cc c}
\toprule
\multirow{2}{*}{\#Itr} & \multicolumn{2}{c}{Bucketed Normalized EPE ($\downarrow$)} & \multicolumn{1}{c}{Efficiency} \\
\cmidrule(lr){2-3} \cmidrule(lr){4-4}
 & Mean Dynamic & Mean Static & FPS \\
\midrule
1 & \textbf{0.1620} & \textbf{0.0079} & \textbf{17.30}\\
2 & 0.1645 & 0.0079 & 16.66\\
3 & 0.1697 & 0.0081 & 15.99\\
4 & 0.1725 & 0.0096 & 14.90\\
5 & 0.1655 & 0.0092 & 13.76\\
\bottomrule
\end{tabular}
\label{table:number_of_decoder}
\end{table}
The component analysis in \cref{table:ablation} demonstrates the contribution of each proposed module. 
In method (i), for both mean Dynamic EPE and mean Static EPE metrics, incorporating STDCB alone brings improvements of 5.1\% and 5.4\% respectively, validating its effectiveness in temporal-spatial feature coupling. 
The application of MambaFlow decoder yields improvements of 3.1\% and 1.1\% respectively, highlighting its capability in feature recovery. 
Meanwhile, utilizing Scene-adaptive Loss alone for supervision training improves mean Dynamic EPE and mean Static EPE by 1.7\% and 15.2\% respectively, demonstrating its robust scene adaptation capability. 
In method (ii), when validating pairwise combinations of components, the model's performance shows substantial improvements, particularly with the synergistic effect of STDCB and scene-adaptive loss, which yields improvements of 7.3\% and 13.0\% in mean Dynamic EPE and mean Static EPE respectively. 
By combining all components, the proposed MambaFlow achieves the best performance.

As shown in \cref{table:number_of_decoder}, the results demonstrate that with a single iteration, 
our method achieves an optimal balance between accuracy and computational speed. We also tested the effect of increasing the number of iterations, and the results show that the increasing number of iterations not only fails to improve performance but also leads to slight degradation in computational efficiency. 
Therefore, we adopt single iteration in our overall architecture. However, further investigation into this phenomenon, particularly on larger-scale datasets and diverse point cloud processing tasks, could yield valuable insights into the optimal number of iterations for different applications.

\subsection{Evaluation of Consumption}
As shown in \Cref{table:resource_consumption}, our method achieves superior performance with only 3.1M parameters, representing a 32.6\% reduction in parameter count compared to Flow4D. Instead of using STDB with redundant convolution operations as in Flow4D, STDCB eliminates duplicate feature extraction processes based on our observation and experimental validation that features extracted in earlier stages already contain sufficient information for subsequent fusion. This architectural optimization also improves computational efficiency, enabling our method to achieve 17.30 FPS compared to Flow4D's 14.68 FPS.

In terms of memory usage, our method achieves a better balance between performance and resource efficiency with moderate memory usage of 2.04 GiB. This slight increase in memory overhead is well compensated by the substantial performance improvements. With a more compact model size of 12.598 MB compared to Flow4D's 18.412 MB, MambaFlow is particularly suitable for deployment in resource-constrained scenarios while maintaining state-of-the-art performance.

\begin{table}
\caption{\textbf{Resource Consumption of different methods.}}
\centering
\setlength{\tabcolsep}{0.4em}
\small
\begin{tabular}{l cccc}
\toprule
Methods & FPS & Params(M) & Total Size(MB) & GM(GiB) \\
\midrule
FastFlow3D   & \textbf{20.35}& 6.8& 27.262& 2.37\\
DeFlow  & \underline{19.82}& 6.9& 27.568 & 2.42 \\
Flow4D  & 14.68&\underline{4.6} &\underline{18.412} & \textbf{1.78}\\
MambaFlow & 17.30 & \textbf{3.1} & \textbf{12.598} & \underline{2.04} \\
\bottomrule
\end{tabular}
\label{table:resource_consumption}
\end{table}

\section{CONCLUSIONS}
In this paper, we present MambaFlow, a novel scene flow estimation approach based on SSM. Through deep coupling of temporal and spatial information, our method achieves comprehensive spatio-temporal feature extraction with temporal guidance. Most importantly, we introduce a Mamba-based decoder that enables refined devoxelization by learning distinct voxel-to-point patterns, effectively preserving fine-grained motion details. Extensive experiments demonstrate that our approach achieves state-of-the-art performance on the Argoverse 2 benchmark while maintaining real-time inference capability, validating the effectiveness of SSM for scene flow estimation.

\addtolength{\textheight}{-4cm}

\bibliographystyle{IEEEtran}
\bibliography{IEEEabrv,MambaFlow}

\begin{thebibliography}{10}
\providecommand{\url}[1]{#1}
\csname url@samestyle\endcsname
\providecommand{\newblock}{\relax}
\providecommand{\bibinfo}[2]{#2}
\providecommand{\BIBentrySTDinterwordspacing}{\spaceskip=0pt\relax}
\providecommand{\BIBentryALTinterwordstretchfactor}{4}
\providecommand{\BIBentryALTinterwordspacing}{\spaceskip=\fontdimen2\font plus
\BIBentryALTinterwordstretchfactor\fontdimen3\font minus \fontdimen4\font\relax}
\providecommand{\BIBforeignlanguage}[2]{{%
\expandafter\ifx\csname l@#1\endcsname\relax
\typeout{** WARNING: IEEEtran.bst: No hyphenation pattern has been}%
\typeout{** loaded for the language `#1'. Using the pattern for}%
\typeout{** the default language instead.}%
\else
\language=\csname l@#1\endcsname
\fi
#2}}
\providecommand{\BIBdecl}{\relax}
\BIBdecl

\bibitem{10258386}
Z.~Yi, H.~Shi, K.~Yang, Q.~Jiang, Y.~Ye \emph{et~al.}, ``Focusflow: Boosting key-points optical flow estimation for autonomous driving,'' \emph{IEEE Transactions on Intelligent Vehicles}, vol.~9, no.~1, pp. 2794--2807, 2024.

\bibitem{10458341}
H.~Liu, Z.~Huang, X.~Mo, and C.~Lv, ``Augmenting reinforcement learning with transformer-based scene representation learning for decision-making of autonomous driving,'' \emph{IEEE Transactions on Intelligent Vehicles}, vol.~9, no.~3, pp. 4405--4421, 2024.

\bibitem{10570336}
H.~Shi, Q.~Jiang, K.~Yang, X.~Yin, H.~Ni \emph{et~al.}, ``Beyond the field-of-view: Enhancing scene visibility and perception with clip-recurrent transformer,'' \emph{IEEE Transactions on Intelligent Vehicles}, pp. 1--16, 2024.

\bibitem{10265176}
H.~Meng, C.~Li, G.~Chen, L.~Chen, and A.~Knoll, ``Efficient 3d object detection based on pseudo-lidar representation,'' \emph{IEEE Transactions on Intelligent Vehicles}, vol.~9, no.~1, pp. 1953--1964, 2024.

\bibitem{10680561}
Y.~Feng, Z.~Guo, Q.~Chen, and R.~Fan, ``Scipad: Incorporating spatial clues into unsupervised pose-depth joint learning,'' \emph{IEEE Transactions on Intelligent Vehicles}, pp. 1--11, 2024.

\bibitem{10640268}
S.~Kim, C.~Kim, and K.~Jo, ``Awv-mos-lio: Adaptive window visibility based moving object segmentation with lidar inertial odometry,'' \emph{IEEE Transactions on Intelligent Vehicles}, pp. 1--16, 2024.

\bibitem{li2021neural}
X.~Li, J.~Kaesemodel~Pontes, and S.~Lucey, ``Neural scene flow prior,'' \emph{Advances in Neural Information Processing Systems}, vol.~34, pp. 7838--7851, 2021.

\bibitem{li2023fast}
X.~Li, J.~Zheng, F.~Ferroni, J.~K. Pontes, and S.~Lucey, ``Fast neural scene flow,'' in \emph{Proceedings of the IEEE/CVF International Conference on Computer Vision}, 2023, pp. 9878--9890.

\bibitem{lin2024icp}
Y.~Lin and H.~Caesar, ``Icp-flow: Lidar scene flow estimation with icp,'' in \emph{Proceedings of the IEEE/CVF Conference on Computer Vision and Pattern Recognition}, 2024, pp. 15\,501--15\,511.

\bibitem{zhang2024seflow}
Q.~Zhang, Y.~Yang, P.~Li, O.~Andersson, and P.~Jensfelt, ``{SeFlow}: A self-supervised scene flow method in autonomous driving,'' in \emph{European Conference on Computer Vision (ECCV)}.\hskip 1em plus 0.5em minus 0.4em\relax Springer, 2024, p. 353–369.

\bibitem{Jund_Sweeney_Abdo_Chen_Shlens_2022}
\BIBentryALTinterwordspacing
P.~Jund, C.~Sweeney, N.~Abdo, Z.~Chen, and J.~Shlens, ``\BIBforeignlanguage{en-US}{Scalable scene flow from point clouds in the real world},'' \emph{\BIBforeignlanguage{en-US}{IEEE Robotics and Automation Letters}}, p. 1589–1596, Apr 2022. [Online]. Available: \url{http://dx.doi.org/10.1109/lra.2021.3139542}
\BIBentrySTDinterwordspacing

\bibitem{zhang2024deflow}
Q.~Zhang, Y.~Yang, H.~Fang, R.~Geng, and P.~Jensfelt, ``{DeFlow}: Decoder of scene flow network in autonomous driving,'' in \emph{2024 IEEE International Conference on Robotics and Automation (ICRA)}, 2024, pp. 2105--2111.

\bibitem{khatri2025can}
I.~Khatri, K.~Vedder, N.~Peri, D.~Ramanan, and J.~Hays, ``I can’t believe it’s not scene flow!'' in \emph{European Conference on Computer Vision}.\hskip 1em plus 0.5em minus 0.4em\relax Springer, 2025, pp. 242--257.

\bibitem{kim2024flow4d}
J.~Kim, J.~Woo, U.~Shin, J.~Oh, and S.~Im, ``Flow4d: Leveraging 4d voxel network for lidar scene flow estimation,'' \emph{arXiv preprint arXiv:2407.07995}, 2024.

\bibitem{wu2024point}
X.~Wu, L.~Jiang, P.-S. Wang, Z.~Liu, X.~Liu \emph{et~al.}, ``Point transformer v3: Simpler faster stronger,'' in \emph{Proceedings of the IEEE/CVF Conference on Computer Vision and Pattern Recognition}, 2024, pp. 4840--4851.

\bibitem{10504607}
J.~Li, Y.~Zhang, P.~Yun, G.~Zhou, Q.~Chen \emph{et~al.}, ``Roadformer: Duplex transformer for rgb-normal semantic road scene parsing,'' \emph{IEEE Transactions on Intelligent Vehicles}, vol.~9, no.~7, pp. 5163--5172, 2024.

\bibitem{ssm2}
A.~Gu, K.~Goel, and C.~R{\'e}, ``Efficiently modeling long sequences with structured state spaces,'' \emph{arXiv preprint arXiv:2111.00396}, 2021.

\bibitem{ssm3}
J.~T. Smith, A.~Warrington, and S.~W. Linderman, ``Simplified state space layers for sequence modeling,'' \emph{arXiv preprint arXiv:2208.04933}, 2022.

\bibitem{ssm5}
D.~Y. Fu, T.~Dao, K.~K. Saab, A.~W. Thomas, A.~Rudra \emph{et~al.}, ``Hungry hungry hippos: Towards language modeling with state space models,'' \emph{arXiv preprint arXiv:2212.14052}, 2022.

\bibitem{mamba3d}
X.~Han, Y.~Tang, Z.~Wang, and X.~Li, ``Mamba3d: Enhancing local features for 3d point cloud analysis via state space model,'' \emph{arXiv preprint arXiv:2404.14966}, 2024.

\bibitem{liang2024pointmamba}
D.~Liang, X.~Zhou, W.~Xu, X.~Zhu, Z.~Zou \emph{et~al.}, ``Pointmamba: A simple state space model for point cloud analysis,'' \emph{arXiv preprint arXiv:2402.10739}, 2024.

\bibitem{mambamos}
\BIBentryALTinterwordspacing
K.~Zeng, H.~Shi, J.~Lin, S.~Li, J.~Cheng \emph{et~al.}, ``Mambamos: Lidar-based 3d moving object segmentation with motion-aware state space model,'' 2024. [Online]. Available: \url{https://arxiv.org/abs/2404.12794}
\BIBentrySTDinterwordspacing

\bibitem{liu2024difflow3d}
J.~Liu, G.~Wang, W.~Ye, C.~Jiang, J.~Han \emph{et~al.}, ``Difflow3d: Toward robust uncertainty-aware scene flow estimation with iterative diffusion-based refinement,'' in \emph{Proceedings of the IEEE/CVF Conference on Computer Vision and Pattern Recognition}, 2024, pp. 15\,109--15\,119.

\bibitem{wang20233d}
Z.~Wang, Y.~Wei, Y.~Rao, J.~Zhou, and J.~Lu, ``3d point-voxel correlation fields for scene flow estimation,'' \emph{IEEE Transactions on Pattern Analysis and Machine Intelligence}, 2023.

\bibitem{Chang_Funkhouser_Guibas_Hanrahan_Huang_Li_Savarese_Savva_Song_Su_etal._2015}
A.~Chang, T.~Funkhouser, L.~Guibas, P.~Hanrahan, Q.~Huang \emph{et~al.}, ``\BIBforeignlanguage{en-US}{Shapenet: An information-rich 3d model repository},'' \emph{\BIBforeignlanguage{en-US}{arXiv: Graphics,arXiv: Graphics}}, Dec 2015.

\bibitem{mayer2016large}
N.~Mayer, E.~Ilg, P.~Hausser, P.~Fischer, D.~Cremers \emph{et~al.}, ``A large dataset to train convolutional networks for disparity, optical flow, and scene flow estimation,'' in \emph{Proceedings of the IEEE conference on computer vision and pattern recognition}, 2016, pp. 4040--4048.

\bibitem{argoverse2}
\BIBentryALTinterwordspacing
B.~Wilson, W.~Qi, T.~Agarwal, J.~Lambert, J.~Singh \emph{et~al.}, ``Argoverse 2: Next generation datasets for self-driving perception and forecasting,'' in \emph{Proceedings of the Neural Information Processing Systems Track on Datasets and Benchmarks}, J.~Vanschoren and S.~Yeung, Eds., vol.~1, 2021. [Online]. Available: \url{https://datasets-benchmarks-proceedings.neurips.cc/paper_files/paper/2021/file/4734ba6f3de83d861c3176a6273cac6d-Paper-round2.pdf}
\BIBentrySTDinterwordspacing

\bibitem{waymo}
P.~Sun, H.~Kretzschmar, X.~Dotiwalla, A.~Chouard, V.~Patnaik \emph{et~al.}, ``Scalability in perception for autonomous driving: Waymo open dataset,'' in \emph{Proceedings of the IEEE/CVF conference on computer vision and pattern recognition}, 2020, pp. 2446--2454.

\bibitem{liu2019flownet3d}
X.~Liu, C.~R. Qi, and L.~J. Guibas, ``Flownet3d: Learning scene flow in 3d point clouds,'' in \emph{Proceedings of the IEEE/CVF conference on computer vision and pattern recognition}, 2019, pp. 529--537.

\bibitem{vedder2022sparse}
K.~Vedder and E.~Eaton, ``Sparse pointpillars: Maintaining and exploiting input sparsity to improve runtime on embedded systems,'' in \emph{2022 IEEE/RSJ International Conference on Intelligent Robots and Systems (IROS)}.\hskip 1em plus 0.5em minus 0.4em\relax IEEE, 2022, pp. 2025--2031.

\bibitem{vedder2024scene}
K.~Vedder, N.~Peri, I.~Khatri, S.~Li, E.~Eaton \emph{et~al.}, ``Scene flow as a partial differential equation,'' \emph{arXiv preprint arXiv:2410.02031}, 2024.

\bibitem{gu2023mamba}
A.~Gu and T.~Dao, ``Mamba: Linear-time sequence modeling with selective state spaces,'' \emph{arXiv preprint arXiv:2312.00752}, 2023.

\bibitem{visionmamba}
L.~Zhu, B.~Liao, Q.~Zhang, X.~Wang, W.~Liu \emph{et~al.}, ``Vision mamba: Efficient visual representation learning with bidirectional state space model,'' \emph{arXiv preprint arXiv:2401.09417}, 2024.

\bibitem{efficientvmamba}
X.~Pei, T.~Huang, and C.~Xu, ``Efficientvmamba: Atrous selective scan for light weight visual mamba,'' \emph{arXiv preprint arXiv:2403.09977}, 2024.

\end{thebibliography}

\begin{IEEEbiographynophoto}
{Jiehao Luo}(Student Member, IEEE) is currently conducting research under the supervision of Xiaoyu Tang at the School of Data Science and Engineering, and Xingzhi College, South China Normal University. His research focuses on computer vision and robotic perception.
\end{IEEEbiographynophoto}

\begin{IEEEbiographynophoto}
{Jintao Cheng} received his bachelor's degree from the School of Physics and Telecommunications Engineering, South China Normal University, in 2021. His research includes computer vision, SLAM, and deep learning.
\end{IEEEbiographynophoto}

\begin{IEEEbiographynophoto}{Xiaoyu Tang}
(Member, IEEE) received a B.S. degree from South China Normal University, Guangzhou, China, in 2003 and an M.S. degree from Sun Yat-sen University, Guangzhou, China, in 2011. He is currently pursuing a Ph.D. degree with South China Normal University. Mr. Tang works as an associate professor, master supervisor, and deputy dean at Xingzhi College, South China Normal University. His research interests include image processing and intelligent control, artificial intelligence, the Internet of Things, and educational informatization.
\end{IEEEbiographynophoto}

\begin{IEEEbiographynophoto}{Qingwen Zhang}
(Student Member, IEEE)
received the M.Phil. degree from The Hong Kong University of Science and Technology in 2022. She is currently a Ph.D student at the KTH Royal Institute of Technology. Her research interests include dynamic awareness in point clouds.
\end{IEEEbiographynophoto}

\begin{IEEEbiographynophoto}{Bohuan Xue} received the B.Eng. degree in computer science and technology from College of Mobile Telecommunications, Chongqing University of Posts and and Telecom, Chongqing, China, in 2018, and the Ph.D. degree from the Department of Computer Science and Engineering, the Hong Kong University of Science and Technology, Hong Kong, China, in 2024. He is currently working as research assistant in School of Data Science and Engineering, and Xingzhi College, South China Normal University.
\end{IEEEbiographynophoto}

\begin{IEEEbiographynophoto}{Rui Fan}(Senior Member, IEEE) received the B.Eng. degree in Automation from the Harbin Institute of Technology in 2015 and the Ph.D. degree in Electrical and Electronic Engineering from the University of Bristol in 2018. He worked as a Research Associate at the Hong Kong University of Science and Technology from 2018 to 2020 and a Postdoctoral Scholar-Employee at the University of California San Diego between 2020 and 2021. He began his faculty career as a Full Research Professor with the College of Electronics \& Information Engineering at Tongji University in 2021, and was then promoted to a Full Professor in the same college, as well as at the Shanghai Research Institute for Intelligent Autonomous Systems in 2022. 

Prof. Fan served as an associate editor for ICRA'23 and IROS'23/24, an area chair for ICIP'24, and a senior program committee member for AAAI'23/24/25. He is the general chair of the AVVision community and organized several impactful workshops and special sessions in conjunction with WACV'21, ICIP'21/22/23, ICCV'21, and ECCV'22. He was honored by being included in the Stanford University List of Top 2\% Scientists Worldwide between 2022 and 2024, recognized on the Forbes China List of 100 Outstanding Overseas Returnees in 2023, and acknowledged as one of Xiaomi Young Talents in 2023. His research interests include computer vision, deep learning, and robotics, with a specific focus on humanoid visual perception under the two-streams hypothesis.
\end{IEEEbiographynophoto}

\vfill

\end{document}